\documentclass[ctexart, utf8, 11pt]{article}
\usepackage{nodalida2021}
\usepackage{times}
\usepackage{soul}
\usepackage{epstopdf}
\usepackage{booktabs}
\usepackage{listings}
\usepackage{breqn}
\usepackage{amsmath}
\usepackage{comment}
\usepackage{graphicx}
\usepackage{CJKutf8}  % added by Aaron
\usepackage[utf8]{inputenc}
\usepackage[T1]{fontenc}
\usepackage{graphicx,amsmath,amsfonts,amssymb,amscd,amsthm}
\usepackage{url}
\usepackage{hyperref} %added by Aaron
\usepackage{latexsym}
\usepackage{pgfplots}
\pgfplotsset{compat=1.14}
\usepackage{pgfplotstable}

\definecolor{aqua}{rgb}{0.0, 1.0, 1.0}

\newlength\graphheight
\newlength\graphwidth
\pgfplotsset {
graph/.style={
height=\graphheight,
width=\graphwidth,
legend style={
    draw=none,
    fill=none,
    font=\footnotesize,
},
ticklabel style={font=\scriptsize, /pgf/number format/fixed},
scaled y ticks = false,
scaled x ticks = false,
xlabel near ticks,
ylabel near ticks,
},
graphright/.style={
graph,
yticklabel pos=right,
},
}

 \aclfinalcopy % Uncomment this line for the final submission
\title{Meta-Evaluation of Translation Evaluation Methods: a systematic up-to-date overview}

\author{ Lifeng Han \& Serge Gladkoff \\
The University of Manchester, UK \& Leiden University, NL\\
Logrus Global LLC \\
Presented Tutorial at LREC-2022: International Conference on Language Resources and Evaluation \\
Lifeng.han@manchester.ac.uk, l.han@lumc.nl \& serge.gladkoff@logrusglobal.com }
         
\date{}

\begin{document}

\maketitle

\begin{figure}[!t]
\centering
\includegraphics*[height=2.1in,width=3in]{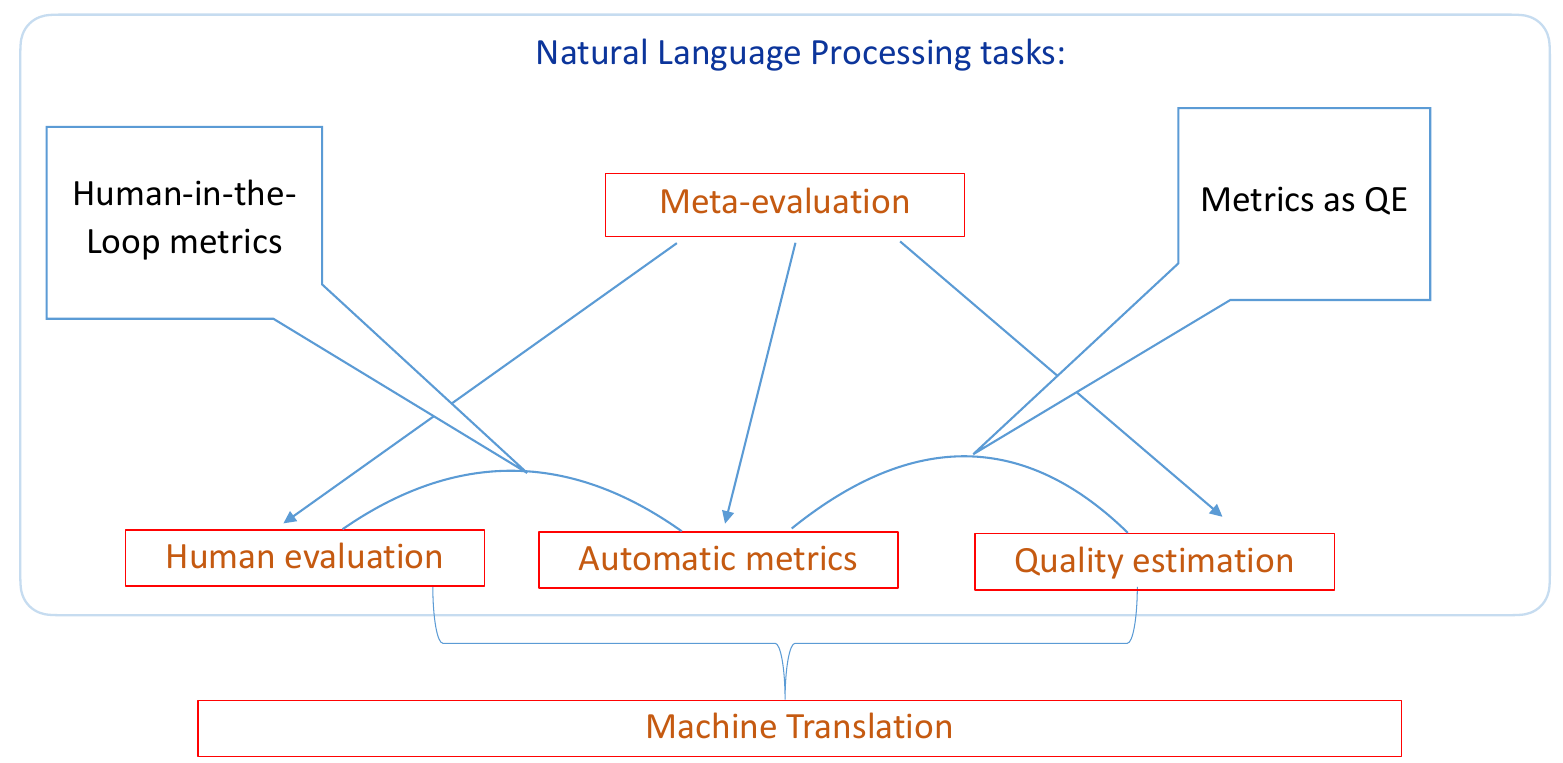}
%[height=3.1in,width=2.6in]
\caption{Meta-eval of MTE: apply to NLP tasks}
\label{fig:meta-eval2nlp}
\end{figure}

\begin{figure}[!t]
\centering
\includegraphics*[height=2.1in,width=3in]{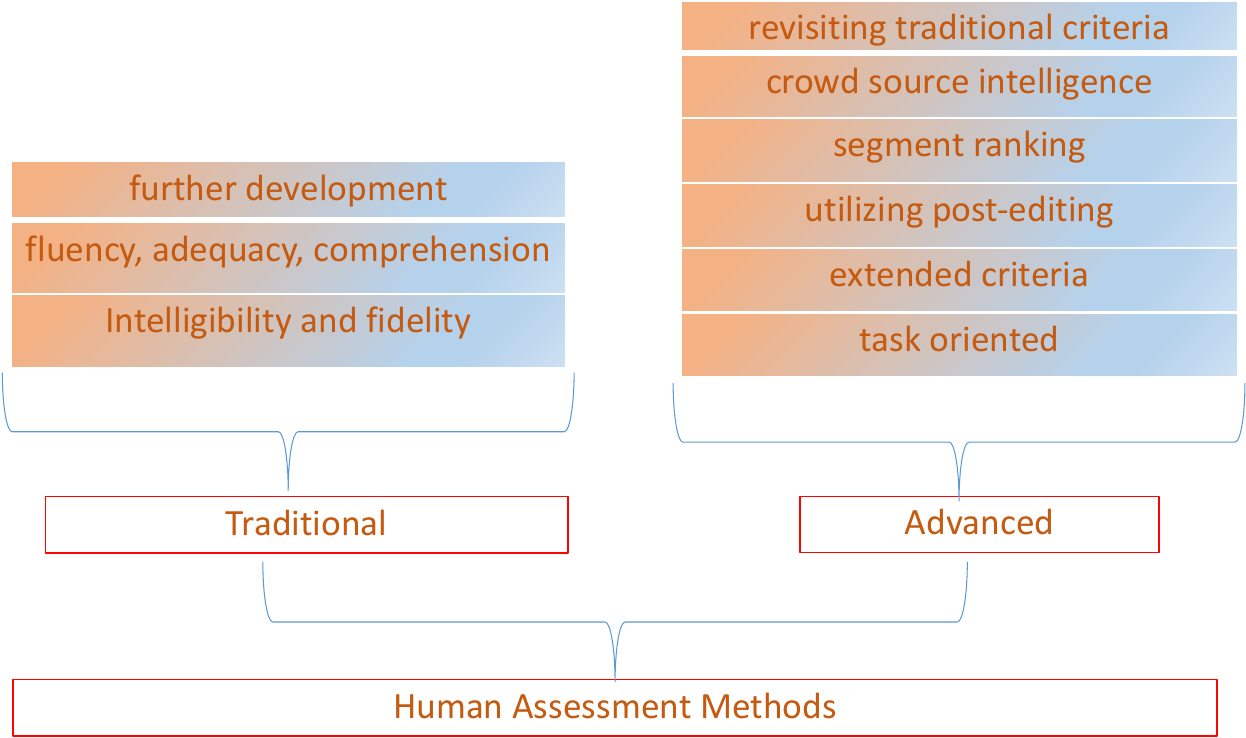}
%[height=3.1in,width=2.6in]
\caption{Human Quality Assessment Methods}
\label{fig:humanMTEfig}
\end{figure}

\begin{figure}[!t]
\centering
\includegraphics*[height=2.5in,width=3in]{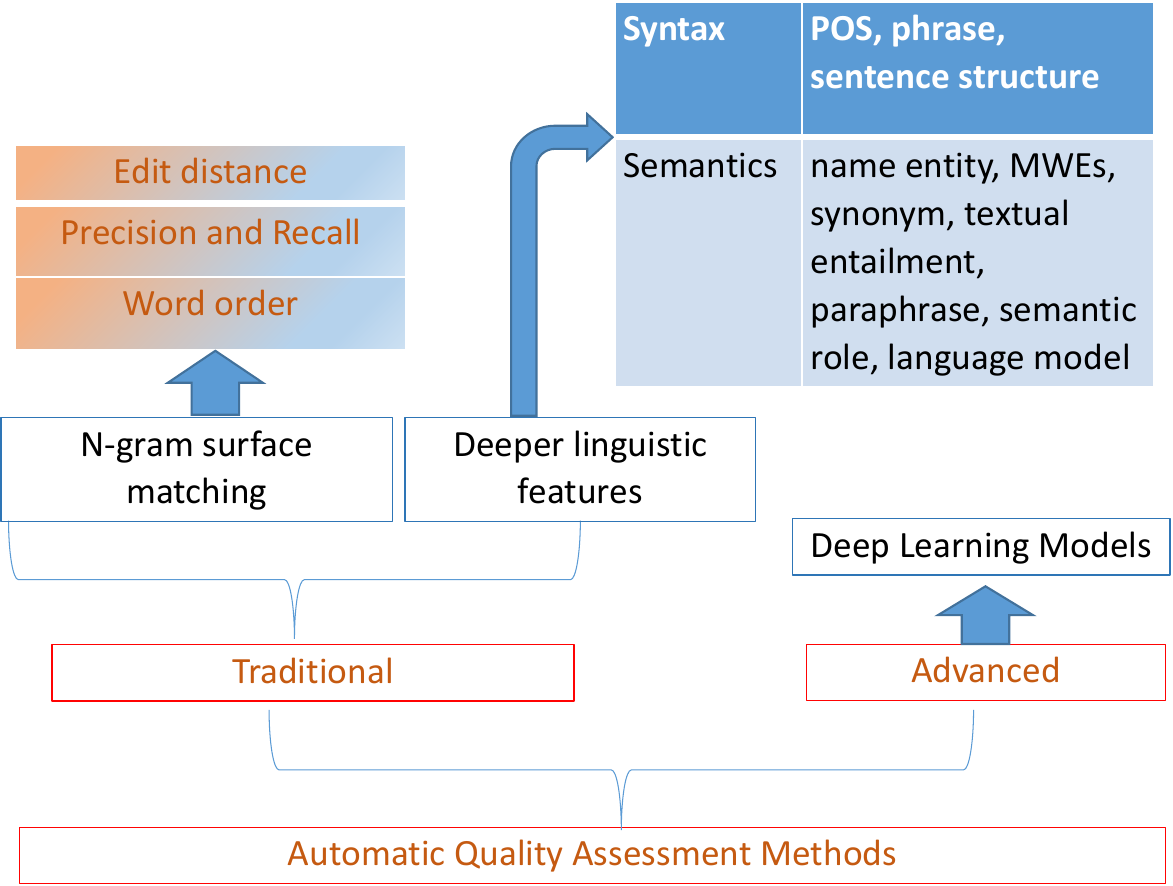}
%[height=3.1in,width=2.6in]
\caption{Automatic Quality Assessment Methods}
\label{fig:automaticMTEfig}
\end{figure}

\begin{abstract}
Starting from 1950s, Machine Translation (MT) was challenged from different scientific solutions which included rule-based methods,  example-based and statistical models (SMT), to hybrid models, and very recent years the neural models (NMT). 
While NMT has achieved a huge quality improvement in comparison to conventional methodologies, by taking advantages of huge amount of parallel corpora available from internet and the recently developed super computational power support with an acceptable cost, it struggles to achieve real human parity in many domains and most language pairs, if not all of them.
Alongside the long road of MT research and development, quality evaluation metrics played very important roles in MT advancement and evolution.
In this tutorial, we overview the traditional human judgement criteria, automatic evaluation metrics, unsupervised quality estimation models, as well as the meta-evaluation of the evaluation methods. Among these, we will also cover the very recent work in the MT evaluation (MTE) fields taking advantages of large size of pre-trained language models for automatic metric customisation towards exactly deployed language pairs and domains. In addition, we also introduce the statistical confidence estimation regarding sample size needed for human evaluation in real practice simulation. Full tutorial material is \textbf{available} to download at \url{https://github.com/poethan/LREC22_MetaEval_Tutorial}.


\end{abstract}

\section*{Structure of the Tutorial}
In this tutorial, we firstly  briefly 
introduce the MT development history, rooted from \textit{language and machines} scientific research field among others at the birth of artificial intelligence (AI) at 1950s, its research paradigms from rule-based to statistical and neural evolution, and 
during this long road, how MT evaluation (MTE) has played its crucial role in advancing the MT technology development. 
This includes conceptual knowledge of earliest manual judgements as golden standard, and lately developed automatic metrics which have been used to evaluate MT algorithms with a low cost and high efficiency, as well as being deployed to optimize the MT model parameters towards better performance. 
Then it comes to the introduction of unsupervised quality estimation models that do not rely on human offered reference translations to evaluate the MT output quality, which matches the practical situations when reference translations are often not available.
Subsequently, we present the concept of meta-eval, the evaluation of evaluation methods (\textit{ref} Fig. \ref{fig:meta-eval2nlp}).

After these brief overview of the overall topic, we come to the detailed and structured categorization of each of these subjects: a) human judgements, b) automatic metrics, c) quality estimation, and d) meta-evaluation.

Human Assessment Methods (HAMs), as in Fig. \ref{fig:humanMTEfig}, are classified into  two different branches involving traditional and advanced categories, which the first includes intelligibility and fidelity, fluency, adequacy, comprehension, and further development, and the second includes task oriented, extended criteria, utilizing post-editing, segment ranking, crowd source intelligence, and some recent work revisiting traditional criteria.
The surveyed research work on HAMs dating back to MT stating point and up to current state-of-the-art.

The automatic evaluation metrics (AEMs) started their development from the late 1990s when statistical MT (SMT) was getting popular and making progress regarding translation output quality. SMT systems were quite often updated using newly developed algorithms and model features and these need very efficient automatic evaluation with a low cost and repeatable performance, which the conventional human input based methods can not afford. 
The AEMs witnessed the methodological changing from simple n-gram based word matching, to deeper linguistic features integration, to nowadays deep learning (DL) based neural network models including the usage of pre-trained language models, Fig. \ref{fig:automaticMTEfig}. 
We classify the first two methodologies into traditional and the DL models into advanced one, of which the n-gram matching category covers editing-distance, precision, recall, and word order features, while the linguistic features category includes both syntax and semantics.
The syntactic features contain Part-of-Speech (POS), Phrase and sentence structures information, and the semantic features include even broader areas such as named entities, idiomatic multi-word expressions (MWEs), synonyms, textual entailment, paraphrase, semantic role labelling, and language models, etc.
This part of tutorial covers most of the metrics developed since 1990s to date (\textit{ref.} Appendix).

Following the development of automatic metrics, we introduce the quality estimation  (QE) research and the evaluation methods defined for QE, which started from 2012 to date, as an affiliated shared task with the annual workshop on  SMT (WMT). The QE models try to extract the knowledge from source and target sentences via feature engineering instead of using reference translations.
The evaluation methods for QE include DeltaAvg, MAE, and RMSE, etc. that we will explain at length for word/token level and sentence level translation output estimation. The word and token level QE includes functions of ``keep, delete, or replacement'', and the sentence level QE is expected to rank several candidate MT outputs according to their quality, translated from the same source by different models and systems.

Looking back to the overall structure of this tutorial, as in Fig. \ref{fig:meta-eval2nlp}, after the three evaluation paradigms, from human evaluation to metrics and QEs, we present the meta-evaluation of evaluation methods. This includes the statistical significance testing, confidence intervals for sample size simulation, inter and intra-agreement level from human judgment, correlation coefficient between automatic metric and human judgment at both system and segment-level, and metrics comparison methods.
Meta-evaluation places an important role in validating the previous mentioned evaluation methodologies and models.

Finally, we summarize the current issues in MT evaluation, with a discussion and perspectives including: 1) the high-cost in human professional evaluation and the credibility of automatic metrics, 2) the inaccuracies caused by crowd-sourced human evaluation setting from dominant WMT workshops, 3) human-in-the-loop half automatic metrics possibility, 4) recent trend on metrics as quality estimation models, 5) QE in practical application such as for language service providers (LSPs), 6) this meta-eval framework application to general NLP evaluation tasks  (Fig. \ref{fig:meta-eval2nlp}).

\section*{ Lecture List: Motivations}

\begin{itemize}
    \item (M)TE as key factor of MT
    \item HumanEval issues
    \item AutoEval issues
    \item MTEval influence in NLP
    \item Earlier survey and overviews
    \item Our own survey and research papers
    \item Our goals of this tutorial
\end{itemize}

\section*{ Lecture List: HumanEval}

\begin{itemize}
    \item Backgrounds on Eval origins
    \item Intelligibility and Fidelity
    \item Fluency, Adequacy, Comprehension
    \item Further development of these
    \item Task oriented eval
    \item Post-editing based eval
    \item segment ranking from WMT
    \item Crowd source intelligence and DA
    \item Human parity false claims
    \item Multi-word Expressions in HumanEval
    \item MQM vs TAUS DQF (MQM2.0)
    \item Our Human-centric HOPE and demo
    \item Document level eval
    
\end{itemize}

\section*{ Lecture List: AutoEval}

\begin{itemize}
    \item n-gram surface matching
    \item Word-order
    \item Precision and Recall
    \item Levenshtein Distance
    \item Linguistic Features
    \item Syntactic features
    \item POS, phrase, sentence
    \item Semantic features
    \item NE, MWEs, synonym, textual entailment, paraphrase, semantic role, LM
    \item Deep Learning based 
    
    \item Model distillation towards DL/Human
    \item Our work on AutoEval metric using n-gram, syntax, semantics (LEPOR, nLEPOR, hLEPOR, HPPR)
    \item Our work on Model Distillation (cushLEPOR)
    
    \item Quality Estimation (our work using CRFs, SVM, NB)
    
\end{itemize}

\section*{ Lecture List: Meta-Eval}

\begin{itemize}
    \item HumanEval agreement
    \item Correlating AutoEval to HumanEval
    \item Pearson, Spearman, Kendall Tau
    \item Meta-eval example from Google AI
    \item Meta-eval example: eval flaws
    \item Meta-eval example: eval setting
    \item Confidence Intervals and Sample Size (our Monte
Carlo Simulation)

\end{itemize}

\section*{ Lecture List: Ending}

\begin{itemize}
    \item Discussion and Summary
    \item Future research directions
    \item Conclusion
    \item References, Platforms, Tools

\end{itemize}

\section*{Acknowledgments}
The authors thank anonymous reviewers for valuable comments and suggestions on how to improve the tutorial. 
Lifeng Han thanks the support from Prof Goran Nenadic and The University of Manchester. The ADAPT Centre for Digital Content Technology is funded under the SFI Research Centres Programme (Grant 13/RC/2106) and is co-funded under the European Regional Development Fund.

\bibliographystyle{acl_natbib}
\bibliography{nodalida2021}

\appendix

\section*{Selected References}

Selected references for Human Assessment Methods (HAMs), %human evaluation, automatic metrics, QE, and meta-eval: 
Automatic Evaluation Methods (AEMs), Quality Estimation Models (QEs), Meta-eval, and overview are listed below.

\begin{itemize}
\item HAMs \cite{google2021human_evaluation_TQA,NLEDA2016,L_ubli_2020_human_parity,zouhar2021neural_emnlp21_nmt_ped,han-etal-2020-alphamwe,gladkoff2021hope}
\item  AEMs \cite{SnoverDorrSchwartzMicciulla2006,wang2016character,HanWongChao2012,han2013language,han2021cushlepor,wmt17metrics,HanWongChaoHeLiZhu2013}
\item QEs \cite{barrault-etal-wmt2019-findings,specia-etal-2013-quest,fomicheva-etal-2020TACL_QE_unsupervised}
\item Meta-eval \cite{Kendall1938,KendallGibbons1990,barrault-etal-wmt2020_findings,Pearson1900,DBLP:conf/naacl/GrahamBM15,koehn2009statistical,marie-etal-2021-scientific,gladkoff2021measuring}
\item Overview \cite{han-etal-2021-TQA,han2022thesis,han2014lepor,han2016MTE_survey,Han2018iprc} and \cite{han2022overview} in Chinese
%\item 
\end{itemize}

% to add back \cite{HanJonesSmeatonBolzoni2021decomposition4mt_MWE,han-etal-2021-translation}

\end{document}